\documentclass[letterpaper, 10 pt, conference]{ieeeconf}  % Comment this line out if you need a4paper
\usepackage[letterpaper, left=0.75in, right=0.75in, bottom=0.8in, top=0.75in]{geometry}

\IEEEoverridecommandlockouts                              % This command is only needed if 
\usepackage{graphics} % for pdf, bitmapped graphics files
\usepackage{subfigure} %for adding sub-figures 
\usepackage[caption=false,font=footnotesize]{subfig}
\usepackage{epsfig} % for postscript graphics files
\usepackage{mathptmx} % assumes new font selection scheme installed
\usepackage{times} % assumes new font selection scheme installed
\usepackage{amsmath} % assumes amsmath package installed
\usepackage{amssymb}  % assumes amsmath package installed
\usepackage{xcolor}
\usepackage{algorithm}
\usepackage{algpseudocode}
\usepackage{cite} 
\usepackage{import}
\usepackage{url}

\newcommand{\figref}[1]{Fig.~\ref{#1}}

\DeclareMathOperator{\arctantwo}{arctan2}
\newcounter{inlineenum}
\renewcommand{\theinlineenum}{\alph{inlineenum}}
\newenvironment{inlineenum}
  {\unskip\ignorespaces\setcounter{inlineenum}{0}%
   \renewcommand{\item}{\refstepcounter{inlineenum}{({\theinlineenum})~}}}
  {\ignorespacesafterend}

\title{\LARGE \bf
Autonomous flight through cluttered outdoor environments \\ using a memoryless planner
}

\author{Junseok Lee$^{1}$$^{\dagger}$, Xiangyu Wu$^{1}$$^{\dagger}$, Seung Jae Lee$^{2}$, and Mark W. Mueller$^{1}$% <-this % stops a space
\thanks{$\dagger$Junseok Lee and Xiangyu Wu contributed equally to this article. Names are in alphabetical order.}%
\thanks{$^{1}$Junseok Lee, Xiangyu Wu, and Mark W. Mueller are with the High Performance Robotics Lab (HiPeRLab) at the Department of Mechanical Engineering, UC Berkeley, CA 94720, USA
        {\tt\small \{junseok\_lee,wuxiangyu,mwm\}@berkeley.edu}}%
\thanks{$^{2}$Seung Jae Lee is with the Automation and Systems Research Institute (ASRI), Seoul National University,
        Seoul, Republic of Korea
        {\tt\small sjlazza@snu.ac.kr}}%
}

\begin{document}

\maketitle
\thispagestyle{empty}
\pagestyle{empty}

%%%%%%%%%%%%%%%%%%%%%%%%%%%%%%%%%%%%%%%%%%%%%%%%%%%%%%%%%%%%%%%%%%%%%%%%%%%%%%%%
\begin{abstract}
This paper introduces a collision avoidance system for navigating a multicopter in cluttered outdoor environments based on the recent memory-less motion planner, rectangular pyramid partitioning using integrated depth sensors (RAPPIDS).
The RAPPIDS motion planner generates collision-free flight trajectories at high speed with low computational cost using only the latest depth image.
In this work we extend it to improve the performance of the planner by taking the following issues into account.
\begin{inlineenum}
  \item Changes in the dynamic characteristics of the multicopter that occur during flight, such as changes in motor input/output characteristics due to battery voltage drop.
  \item The noise of the flight sensor, which can cause unwanted control input components.
  \item Planner utility function which may not be suitable for the cluttered environment.
\end{inlineenum}
Therefore, in this paper we introduce solutions to each of the above problems and propose a system for the successful operation of the RAPPIDS planner in an outdoor cluttered flight environment.
At the end of the paper, we validate the proposed method's effectiveness by presenting the flight experiment results in a forest environment. A video can be found at \url{www.youtube.com/channel/UCK-gErmvZlBODN5gQpNcpsg}
\end{abstract}

%%%%%%%%%%%%%%%%%%%%%%%%%%%%%%%%%%%%%%%%%%%%%%%%%%%%%%%%%%%%%%%%%%%%%%%%%%%%%%%%
\section{INTRODUCTION}
Motion planning algorithms for multicopter unmanned aerial vehicles to fly autonomously to their destination in cluttered environments are in general grouped into two categories.
One is to separately run a path planning algorithm to generate collision-free path as a purely geometrical problem without considering dynamics \cite{pollefeys_obstacle_nodate}, and use a path follower to follow the collision-free path.
Since it does not consider dynamics constraints, collision avoidance can not be guaranteed at high speed since dynamics constraints, such as motor thrust limit, are not considered at the time of planning.
The other approach considers dynamics constraints and directly generates control commands by including obstacle avoidance as a constraint inside an optimization problem rather than separating path planning and tracking, for example, based on the rapidly-exploring random tree star (RRT*) \cite{sakcak_sampling-based_2019}, the nonlinear optimization \cite{spedicato_minimum-time_2018}, and the mixed-integer programming (MIP) \cite{park_homotopy-based_2015}.

We focus on the latter category which considers dynamics in planning as well as collision avoidance.
The collision-free trajectories are often further constrained by minimizing a cost function depending on applications, for example, the minimum time, the minimum energy, and the shortest distance.
Trajectory generation algorithms may be divided into two major types: map-based algorithms and memory-less algorithms \cite{avoidance_algorithms}.

\begin{figure}
	\begin{center}    \includegraphics[width = 0.9 \linewidth]{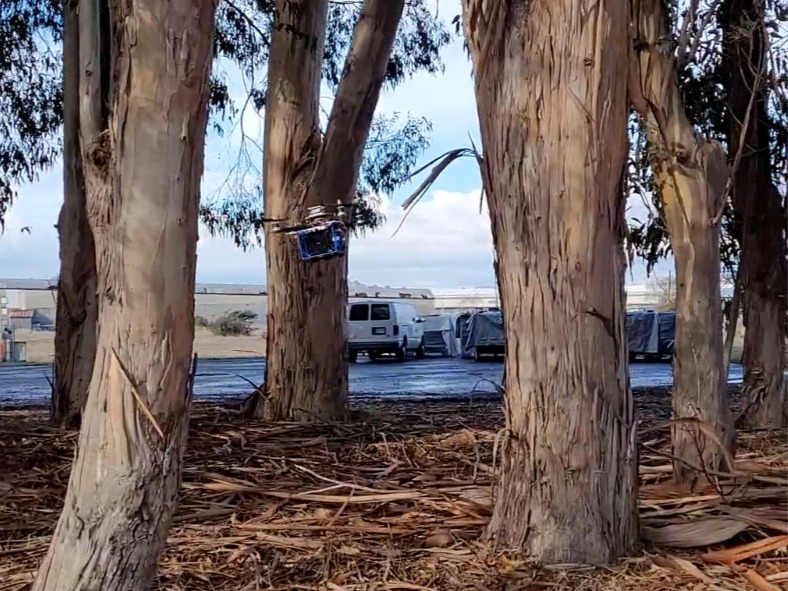}
	\end{center}
	\caption{The vehicle flies autonomously using visual-inertial odometry and the collision-avoidance planner through a forest, avoiding trees.}
	\label{fig:front-page}
\end{figure}
Map-based algorithms are global planning methods of creating collision-free optimal trajectories after a single large map is constructed or while creating a map by fusing all spatial sensor information obtained during flights. For instance, in \cite{map-based_1}, a local map map of the environment is used and a nonconvex, nonlinear optimization problem is solved to get collision-free smooth trajectories. In \cite{map-based_3,map-based_4,map-based_5}  the free-space in the map is represented as multiple convex regions, and an optimization problem is then solved to find a series of trajectories through the free-space. In \cite{B-spline1}, the convex hull property of B-spline trajectories is used to solve for safe and fast trajectories, and the success rate and optimality is improved in the subsequent work of \cite{B-spline2}.
Map-based algorithms have the advantage of optimal trajectory generation because it presupposes that map information is already known when planning.
However, they usually have high computational cost and due to fusing sensor data into the map.

On the other hand, memory-less algorithms use only the most recent sensor information to avoid obstacles, such as using the k-d-tree \cite{memoryless_1,memoryless_2}, and a trajectory library precomputed offline to reduce a significant amount of online computation \cite{memoryless_4}.
Therefore, memory-less algorithms are classified as local planning methods and are advantageous for obstacle avoidance in a dynamic obstacle environment due to low computational cost and high update frequency.
However, there is a disadvantage that it is challenging to find a globally optimal trajectory since global spatial information is not available at the time of planning.

For trajectory generation of a small-size multicopter that requires high-speed maneuver but has a limited payload capacity, memory-less algorithms have great utility due to the following reasons.
First, memory-less algorithms are easy to be implemented in real-time on miniature on-board computers with limited computational resources.
Second, due to the fast trajectory update speed thanks to the low computational cost, memory-less algorithms can cope with rapidly changing surroundings during high-speed flight.
Lastly, the algorithm is less prone to accumulated odometry errors during flights in a cluttered environment because it utilizes only the latest sensor information for the planning. 

Recently, a memory-less planner is proposed using rectangular pyramid partitioning using integrated depth sensors (RAPPIDS) motion planner, which has high computational efficiency for high-speed collision avoidance flights \cite{rappids_nathan}.
Using the RAPPIDS motion planner, the authors were able to achieve high collision avoidance flight performance in cluttered indoor flight environment, using only stand-alone depth images and visual-inertial odometry information processed by an on-board computer mounted on the vehicle.

In this paper, we extend the RAPPIDS planner to operate in a cluttered outdoor off-road environment.
We describe the system's development process and flight results.
In the experiment, the system was able to fly 30 meters in a forest environment, as shown in Fig. \ref{fig:front-page}, with a maximum speed of 2.7 m/s.
The remaining parts of the paper are organized as follows:
in Section II, we outline the principles of the RAPPIDS planner.
Section III introduces modifications to the algorithm to ensure successful outdoor flights.
Finally, in Section IV, the system's performance is demonstrated by introducing the experiment results of obstacle avoidance flights in a forest environment.

%%%%%%%%%%%%%%%%%%%%%%%%%%%%%%%%%%%%%%%%%%%%%%%%%%%%%%%%%%%%%%%%%%%%%%%%%%%%%%%%
\section{RAPPIDS MOTION PLANNING FRAMEWORK} \label{sec:rappids-framework}
\begin{algorithm}[t]
  \centering
  \small
  \caption{\label{alg:constraints-check}  Trajectory Constraint Checks}
  \begin{algorithmic}[1]
    \Require{A sampled candidate trajectory}
    \Procedure{ConstraintsCheck}{}
      \If{lower cost than known}
        % \Comment{Subsection~\ref{ssec:}}
        \If{input feasibility}
        % \Comment{Subsection~\ref{ssec:input-feasibility}}
            \If{velocity admissibility}
            \Comment{Subsection~\ref{ssec:velocity-check}}
                \If{collision-free}
                \Comment{Subsection~\ref{ssec:collision-free}}
                     \State $\textrm{trajectory\_status} \gets \textrm{collision\_free}$
                \Else
                    \State $\textrm{trajectory\_status} \gets \textrm{in\_collision}$
                \EndIf
            \Else
                \State $\textrm{trajectory\_status} \gets \textrm{velocity\_inadmissible}$
            \EndIf
        \Else
            \State $\textrm{trajectory\_status} \gets \textrm{input\_infeasible}$
        \EndIf
      \Else
        \State $\textrm{trajectory\_status} \gets \textrm{higher\_cost}$
      \EndIf
  \EndProcedure
  \end{algorithmic}
\end{algorithm}
In this section, we repeat some details from \cite{rappids_nathan}, and add a velocity limiting check for stable visual-inertial odometry.
Motion primitives are sampled and then %down-selected by 
go through a series of checks to find a trajectory that is minimum-cost, input-feasible, velocity-admissible, and collision-free, as shown in Algorithm~\ref{alg:constraints-check}.
Since the planner has a low computational cost, we plan in a receding horizon fashion every time a new depth image arrives.
This keeps a collision-free trajectory being updated with the latest depth view, and allows avoiding obstacles that are not included in the previous camera view.

%%%%%%%%%%%%%%%%%%%%%%%%%%%%%%%%%%%%%%%%%%%%%%%%%%%%%%
\subsection{Candidate trajectory sampling} 
\label{ssec:traj-sampling}
We sample candidate trajectories first by sampling an endpoint and constructing a polynomial trajectory connecting the current position to the endpoint.
Specifically, we first uniformly sample a 2D point in the pixel coordinates of a depth camera.
We also draw a sampled depth from a uniform distribution, and then back-project the 2D point using the sampled depth to obtain a sampled endpoint $\mathbf{s}_{T}\in\mathbb{R}^{3}$.

Denote $\mathbf{s}(t), \; \dot{\mathbf{s}}(t), \; \textrm{and} \; \ddot{\mathbf{s}}(t) \in\mathbb{R}^{3}$ to be the position, velocity and acceleration of the vehicle in the inertial frame. The candidate motion primitives are described as below.
\begin{equation} \label{eqn:polynomial_trajectory}
    \mathbf{s}(t)=\frac{\mathbf{\alpha}}{120}t^5+\frac{\mathbf{\beta}}{24}t^4+\frac{\mathbf{\gamma}}{6}t^3+\frac{\ddot{\mathbf{s}}(0)}{2}t^2+\dot{\mathbf{s}}(0)t+\mathbf{s}(0),\ t\in[0\ T],
\end{equation}
where $T$ is the trajectory duration, $\mathbf{s}(0)$, $\dot{\mathbf{s}}(0)$, and $\ddot{\mathbf{s}}(0)$ are the initial position, velocity, and acceleration of the vehicle at the time when the trajectory starts, and $\mathbf{\alpha}$, $\mathbf{\beta}$ and $\mathbf{\gamma}$ are coefficients such that $\mathbf{s}(T) = \mathbf{s}_{T}$, and $\dot{\mathbf{s}}(T) = \ddot{\mathbf{s}}(T) = 0$.
This 5th order polynomial corresponds to the minimum-jerk trajectory, which minimizes the average Euclidean norm of jerk over the trajectory duration $T$. The trajectory is smooth and can be checked for collisions efficiently, as shown in \cite{collision-check}.

%%%%%%%%%%%%%%%%%%%%%%%%%%%%%%%%%%%%%%%%%%%%%%%%%%%%%%
\subsection{Velocity constraints for stable visual-inertial odometry \label{ssec:velocity-check}}
We impose velocity constraints to prevent the visual-inertial odometry from losing track in high-speed flights.
The sampled trajectories are filtered out if their maximum velocity exceeds a predefined threshold, $v_{\rm max}$.
For the sake of computational tractability, the constraints are checked for each per-axis velocity using analytical solution for third-order polynomials.
It should be noted that checking the magnitude requires solving higher-order polynomials numerically, because analytical solutions do not exist.
The following equation can be derived by taking the derivative of (\ref{eqn:polynomial_trajectory}),
\begin{equation} \label{eqn:velocity}
    \dot{\mathbf{s}}(t)=
    \frac{\mathbf{\alpha}}{24}t^4+
    \frac{\mathbf{\beta}}{6}t^3+
    \frac{\mathbf{\gamma}}{2}t^2+
    \ddot{\mathbf{s}}(0) t +
    \dot{\mathbf{s}}(0)
\end{equation}
To find its extrema, we compute its derivative and find the zeros as below.
\begin{equation} \label{eqn:velocity_derivative}
    \ddot{\mathbf{s}}(t)=
    \frac{\mathbf{\alpha}}{6}t^3+
    \frac{\mathbf{\beta}}{2}t^2+
    \mathbf{\gamma}t+
    \ddot{\mathbf{s}}(0)
    = 0
\end{equation}
The third-order polynomial can be efficiently solved, and the magnitude of \eqref{eqn:velocity} is evaluated at the roots as well as the boundary, $0$ and $T$.
The procedure is repeated for every axis, and we discard the motion primitive candidate if the speed on any axis exceeds the per-axis velocity limit $v_{\rm max}$.

%%%%%%%%%%%%%%%%%%%%%%%%%%%%%%%%%%%%%%%%%%%%%%%%%%%%%%
\subsection{Collision check: Pyramid method} \label{ssec:collision-free}
\begin{figure}[t]
    \begin{center}
    \includegraphics[width=0.95\columnwidth]{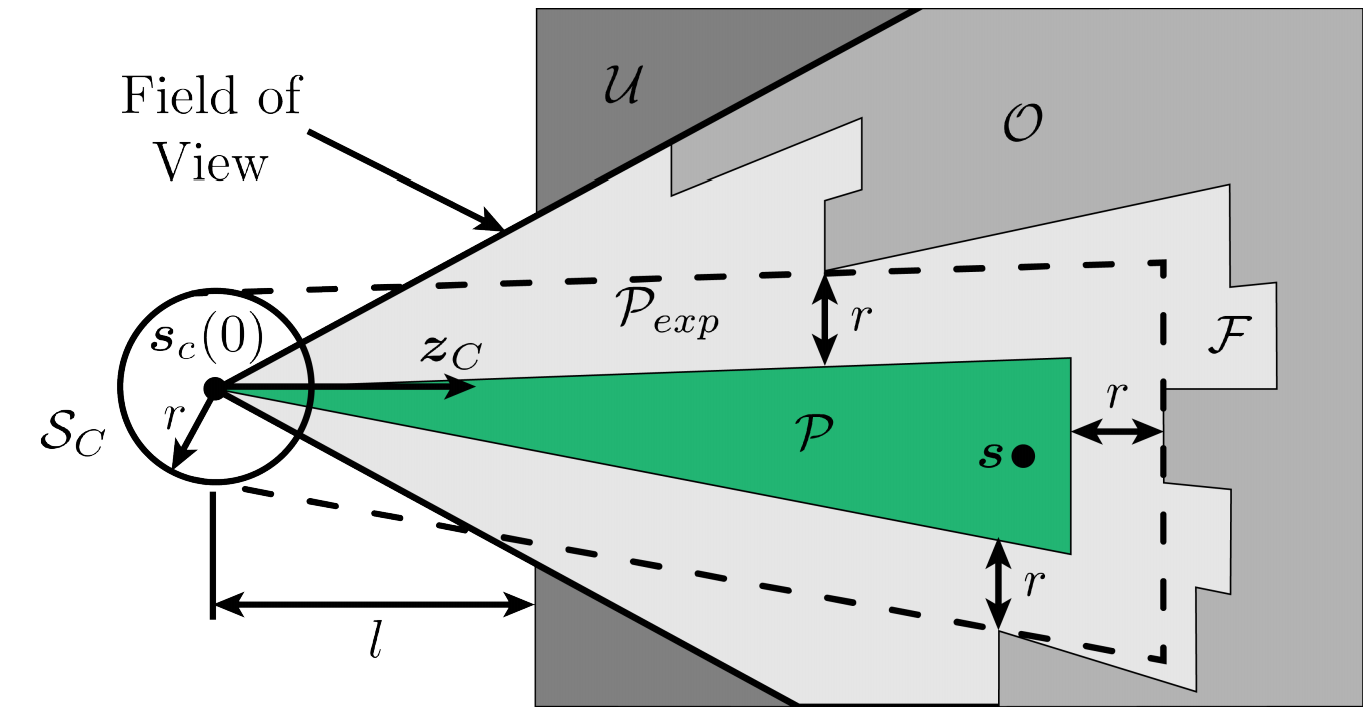}
    \end{center}
    \caption{
    The planner uses a collection of pyramids to represent the free space, which allows simple and fast collision check of a sampled trajectory.
    \textit{The figure is sourced from \cite{rappids_nathan}.}
    }
    \label{gp:depth_image}
\end{figure}
The RAPPIDS planner determines whether the candidate trajectory intrudes the obstacle by pyramid inflation.
Fig. \ref{gp:depth_image} shows the process in which the depth camera on the flying vehicle searches for area $\mathcal{P}$ that guarantees a non-collision path.
First, we define free space $\mathcal{F}$ and occupied space $\mathcal{O}$ based on the depth camera image.
We also treat all spaces outside the field of view that are $l$ distance from the vehicle as occupied spaces to avoid collisions with unrecognized obstacles outside the camera's field of view.
Next, we select the final position $\mathbf{s}(T)$ through random sampling and then search for the nearest depth pixel $p$ from $\mathbf{s}(T)$.
Then, starting at pixel $p$ and reading the surrounding depth pixels in a spiral sequence, we get the largest possible rectangular space $\mathcal{P}_{exp}$ that does not intrude the occupied space $\mathcal{O}$.
Finally, pyramid $\mathcal{P}$ distanced with vehicle radius $r$ is created by shrinking the expanded pyramid $\mathcal{P}_{exp}$.
By checking whether the $\mathbf{s}(t)$ candidate trajectory remaining inside $\mathcal{P}$, we can conclude that the trajectory is collision-free from the detected obstacles.

The algorithm can guarantee zero collision trajectory, because the trajectory generated by the algorithm inherently avoids not only the obstacles recognized by the depth sensor but also the obstacles located in an unobserved ($\mathcal{U})$ or unknown area ($\mathcal{O}$) obscured by the detected obstacles.

%%%%%%%%%%%%%%%%%%%%%%%%%%%%%%%%%%%%%%%%%%%%%%%%%%%%%%

%%%%%%%%%%%%%%%%%%%%%%%%%%%%%%%%%%%%%%%%%%%%%%%%%%%%%%%%%%%%%%%%%%%%%%%%%%%%%%%%
\section{ALGORITHMS FOR OUTDOOR FLIGHT}
In this section, we describe modifications to the motion planner other than the velocity check described in section ~\ref{ssec:velocity-check}, for collision-free flights outdoors to autonomously reach a target waypoint.
\begin{inlineenum}
    \item We sample trajectories with final positions around the center of the view of the depth camera to improve the efficiency of sampling trajectories.
    \item We proposed a new utility function that behaves similarly to the utility function of maximizing the average velocity, but also considers making the vehicle stay around the target.
    \item The vehicle is always yawed towards the goal to dynamically change the view during the flight that can potentially increase the chance of finding a collision-free trajectory compare to the view with a fixed yaw.
    \item The initial acceleration used for sampling trajectories is approximated by the total thrust command divided by the mass instead of using noisy IMU measurements, since the noise in acceleration hampers the planner from finding proper trajectories.
    \item We compensate the thrust change because of battery voltage drop during the flight.
\end{inlineenum}

\begin{figure}
    \centering
    \includegraphics[width=0.95\columnwidth]{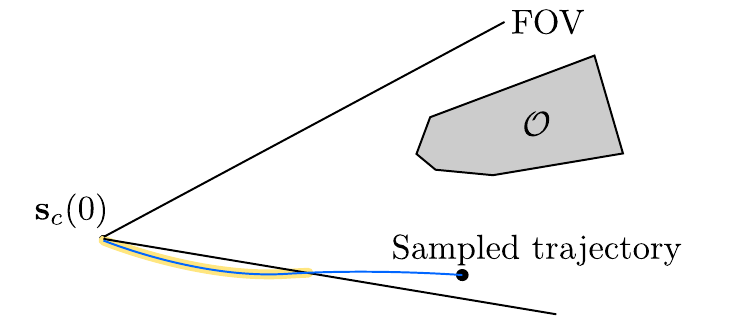}
    \caption{Efficient trajectory sampling in the field of view (FOV) of the depth camera. An occluded space $\mathcal{O}$ at the center makes the planner construct pyramids around the FOV. A sampled trajectory (blue) with an endpoint close to the FOV is likely to have a part that resides outside the field of view (highlighted by yellow), which is classified by the planner as in-collision as space outside of the FOV is considered as occupied. To increase the sampling efficiency, we sample points only between 10\% and 90\% of horizontal and vertical FOV.
    }
    \label{fig:edge_cases}
\end{figure}

\subsection{Sampling efficiency}
\label{subsec:depth_sensor_noise}
As collision checking at the last step of the planner is computationally expensive, it is beneficial to increase the probability of finding a collision-free trajectory from candidate trajectories.
We improve the sampling efficiency by excluding candidate trajectories with high chances of being classified as in collision.
One of the most common cases is when a sampled candidate trajectory has an endpoint around the field of view (FOV), as shown in Fig.~\ref{fig:edge_cases}.
If the endpoint is close to the edges of the FOV, it is likely that some parts of the sampled trajectory fall outside of the FOV, and it is classified by the planner as in-collision because regions outside of the FOV are considered occupied by the planner, as described in section \ref{ssec:collision-free}.
To improve the trajectory sampling efficiency by avoiding those cases, the final trajectory position is sampled between 10\% and 90\% of the field-of-view.

\subsection{Utility function}
\begin{figure}
    \centering
    \includegraphics[width=\columnwidth]{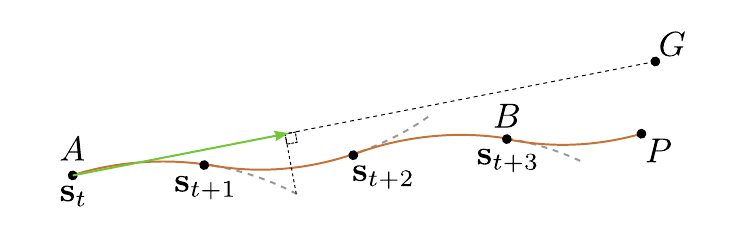}
    \caption{
        A new cost function to maximize average velocity and handle the behavior around the goal point $G$ is proposed.
        Every time a new depth image arrives, the planner attempts to find a new collision-free trajectory using the current estimates $s_t$, and if found, the tracking controller discards the previous trajectory (gray-dashed parts), and starts tracking the new trajectory (brown).
        When far from the goal point, for example point $A$, the utility function is roughly the average velocity measured in direction to the goal (depicted by the green arrow), which still allows lateral motion.
        However, around the goal, such as point $B$, the utility function encourages the planner to generate a new collision-free trajectory with the endpoint around the goal, not beyond the goal.
    }
    \label{fig:cost_function}
\end{figure}

We propose a utility function as below to generate trajectories that not only consider maximizing the average velocity to the goal, but also keep the trajectories' end points around the target.
\begin{equation}
    U(P, \; t) = \frac{\lVert \mathbf{d}_{s_{t}G} \rVert - \lVert \mathbf{d}_{PG}\rVert}{t}, 
\end{equation}
where $\lVert\mathbf{d}_{s_{t}G}\rVert$, $\lVert\mathbf{d}_{PG}\rVert$, and $t$ are the distances between the current position and the goal, the distance between the endpoint of the motion primitives and the goal, and the primitive execution time, respectively.
As shown in Fig.~\ref{fig:cost_function}, when the vehicle is far from the waypoint, the term $\lVert \mathbf{d}_{PG}\rVert$ does not vary much, and hence the utility function is to maximize the average velocity to the waypoint.
Around the waypoint, however, the term $\lVert \mathbf{d}_{PG}\rVert$ plays a role to encourage the planner to choose a trajectory whose final position falls around the goal.

\subsection{Desired yaw angle}
The yaw angle can be arbitrarily chosen while tracking a collision-free trajectory.
We yaw the vehicle always towards the goal, because it is more likely to find a trajectory to the goal when facing towards it.
\begin{align}
    \psi_{c} & = \arctantwo \left(\left[0 \; 1 \; 0\right] (\mathbf{s}_G - \mathbf{s}), \; \left[1 \; 0 \; 0\right] (\mathbf{s}_G - \mathbf{s}) \right)  \\
             & \in \left[-\pi, \pi\right], \nonumber
\end{align}
where $\arctantwo(y, x)$ measures the signed angle between the point $(x, y)$ and the positive x-axis, and $\psi_{c}$, $\mathbf{s}_G$ and $\mathbf{s}$ are the commanded yaw angle, the positions of the goal and the vehicle in the inertial frame, respectively.

\subsection{Acceleration estimation}
Our planner checks collision of a trajectory that is sampled given the current position, velocity, and acceleration (section \ref{ssec:traj-sampling}).
Using acceleration measurements from IMU for the current acceleration results in inaccurate sampled trajectories due to the noise in IMU acceleration measurements.
We instead estimate acceleration using the last commanded thrust as below
\begin{equation}
    \mathbf{\ddot{\mathbf{s}}}\left( 0 \right) = \frac{c\mathbf{z}_B}{m} - \mathbf{g},
\end{equation}
where $c$, $\mathbf{z}_B$, $m$, and $\mathbf{g}$ are the last collective thrust command, body z-axis, mass, and gravitational acceleration.

\subsection{Thrust adaptation}
Once the collision-free trajectory is selected, the cascaded flight controller generates motor speed commands to maneuver the vehicle, as shown in Fig. \ref{fig:block-diagram}.
The commands are then sent to the electric speed controllers (ESCs), where ESCs control each motor's rotation speed based on the predefined protocol in their firmware`  .

\begin{figure}
	\begin{center}    
	\includegraphics[width = 0.95\linewidth]{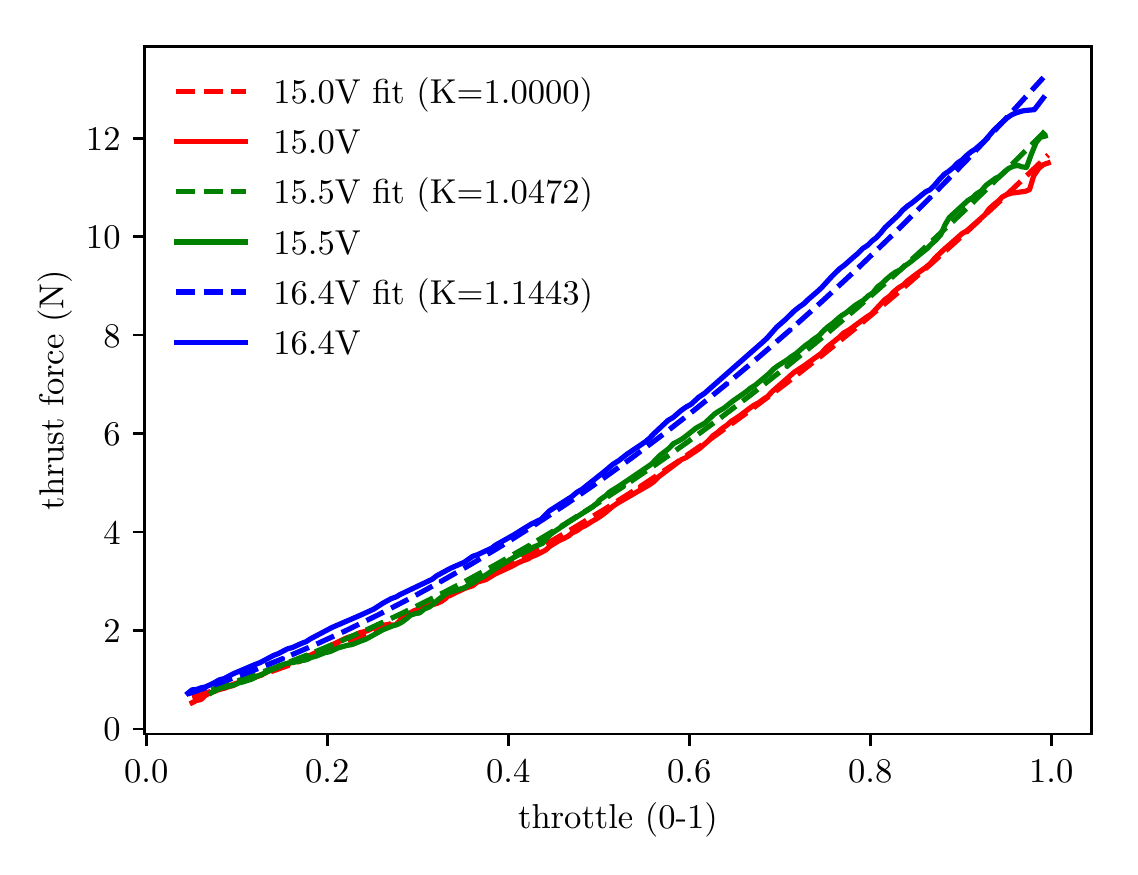}
	\end{center}
	\caption{Thrust force at different thrust commands (solid lines) and second-order fitting results (dotted lines). The command-thrust relationship changes depending on the voltage applied to the propulsion system.}
	\label{fig:thrust_map}
\end{figure}

ESCs can be classified into two groups: open-loop ESCs and closed-loop ESCs.
Unlike closed-loop ESCs that can accurately control motor rotation speed through a feedback control loop, open-loop ESCs used in most multicopters 
control motors to rotate at different speed if the battery's voltage changes, generating different thrust, as shown in Fig. \ref{fig:thrust_map}.
Therefore, to generate the desired thrust force regardless of the voltage level, an on-line thrust model reflecting the current voltage level is required.

\subsubsection{On-line thrust model}
from the data and fitting results in Fig. \ref{fig:thrust_map}, we are able to confirm that the thrust model at each voltage level could be fitted as using the following quadratic function form
\begin{equation}
    f_i(u_i)=K(c_0(u_i+c_1)^2+c_2),
    \label{eq:enhanced_thrust_model}
\end{equation}
where $K$ is the voltage-dependent term, $u_i$ is the thrust command of the $i$-th motor, and $c_{\{0,1,2\}}$ are the nominal model parameters at a voltage level where $K$ is 1.
However, the model includes a fitting error and we additionally structure $K$ as follows to overcome the error
\begin{equation}
    K=k_V k_M,
    \label{eq:K}
\end{equation}
where $k_V$ is the modeled part and $k_M$ is the unmodeled part.
The update rules for $k_V$ and $k_M$ are introduced in the following parts.

\subsubsection{Updating $k_V$ gain}
\begin{figure}
	\begin{center}    \includegraphics[width = 1 \linewidth]{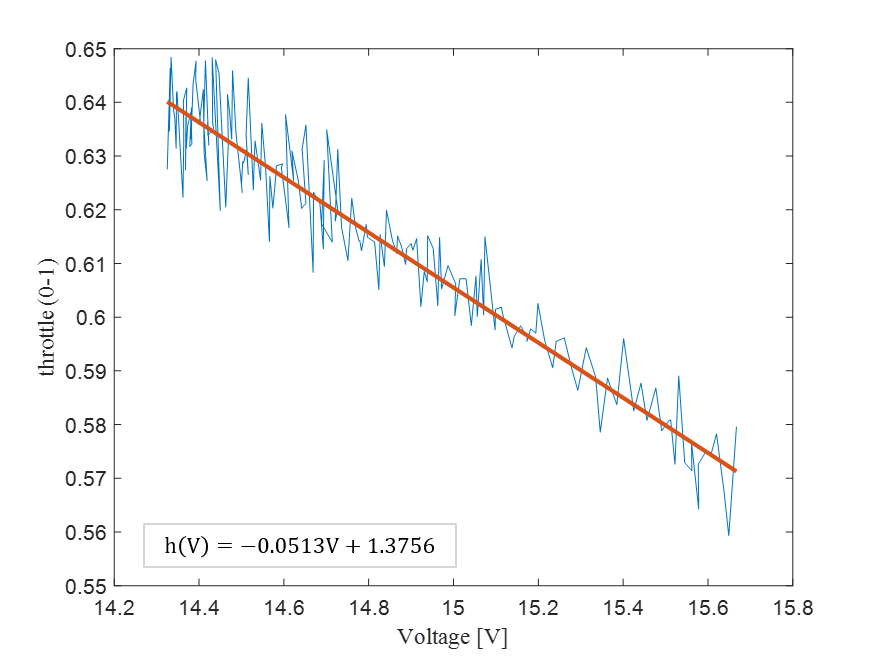}
	\end{center}
	\caption{Hovering command that increases as the voltage decreases. With first-order fitting, we can obtain $h(V)$ function.}
	\label{fig:affinefit}
\end{figure}
fig. \ref{fig:affinefit} shows the hovering thrust throttle command that increases as the voltage level decreases.
From the data, we can derive the affine function $h(V)$ through a first-order fitting.
Then, we compensate for the thrust reduction due to the voltage drop by defining $k_V$ as follows
\begin{equation}
    k_V=\frac{h(V_R)}{h(V)},
    \label{eq:kv}
\end{equation}
where $V_R$ is a reference voltage for which the $c_{\{0,1,2\}}$ values in (\ref{eq:enhanced_thrust_model}) are defined.

\subsubsection{Updating $k_M$ gain}
to compensate for unmodeled errors, we compare the target thrust and the estimated thrust.
The magnitude of actual thrust during flight can be estimated from the acceleration value in the z-direction of the IMU.

First, the translational motion dynamics of a multicopter is described as follows
\begin{equation}
    \label{eq:translationaldynamics}
    m\ddot{\mathbf{s}}=R(\mathbf{q})\mathbf{f}+m\mathbf{g},
\end{equation}
where $\mathbf{q}\in\mathbb{R}^{3}$ is the attitude, $R(\mathbf{q})\in{SO(3)}$ is the rotation matrix that transforms from the body coordinate frame to the world coordinate frame, and $\mathbf{f}=[0\ 0\ \Sigma f_i]^T\in\mathbb{R}^{3}$ is the thrust vector.
Next, the following relationship is established between the IMU acceleration measurement and the actual acceleration
\begin{equation}
    \label{eq:IMUsensor}
    R(\mathbf{q})\mathbf{c}+\mathbf{g}+\boldsymbol{\Delta}_s=\ddot{\mathbf{s}},
\end{equation}
where $\mathbf{c}=[c_x\ c_y\ c_z]^T\in\mathbb{R}^{3}$ is the IMU's acceleration measurement and $\boldsymbol{\Delta}_s$ refers to negligible microterms that occur due to the rotational motion \cite{imu}.

Through (\ref{eq:translationaldynamics}) and (\ref{eq:IMUsensor}), we can bring the following result
\begin{equation}
    mc_z\approx\Sigma f_i,
    \label{eq:thrust_estimation}
\end{equation}
where we can estimate the actual thrust force generated from the vehicle with z-directional IMU acceleration measurements.
Then, we can estimate $k_M$ in real-time as follows
\begin{equation}
    k_M=1+ \int\big(f_i(u_i)-\hat{f}_i\big)dt,\ \hat{f}_i=\frac{mc_z}{n},
    \label{eq:km}
\end{equation}
where $n$ represents the number of thrusters attached to the multicopter.

By updating $k_V$ and $k_M$ gains on-line during the flight through (\ref{eq:kv}) and (\ref{eq:km}), we can calculate control input $u_i$ from the inverse of (\ref{eq:enhanced_thrust_model}) to make each propeller's desired thrust more accurate.

%%%%%%%%%%%%%%%%%%%%%%%%%%%%%%%%%%%%%%%%%%%%%%%%%%%%%%%%%%%%%%%%%%%%%%%%%%%%%%%%
\section{EXPERIMENTAL RESULTS}
\begin{figure}
	\centering
	\subfigure{\hspace*{-0.035\linewidth}\includegraphics[width=0.8\linewidth]{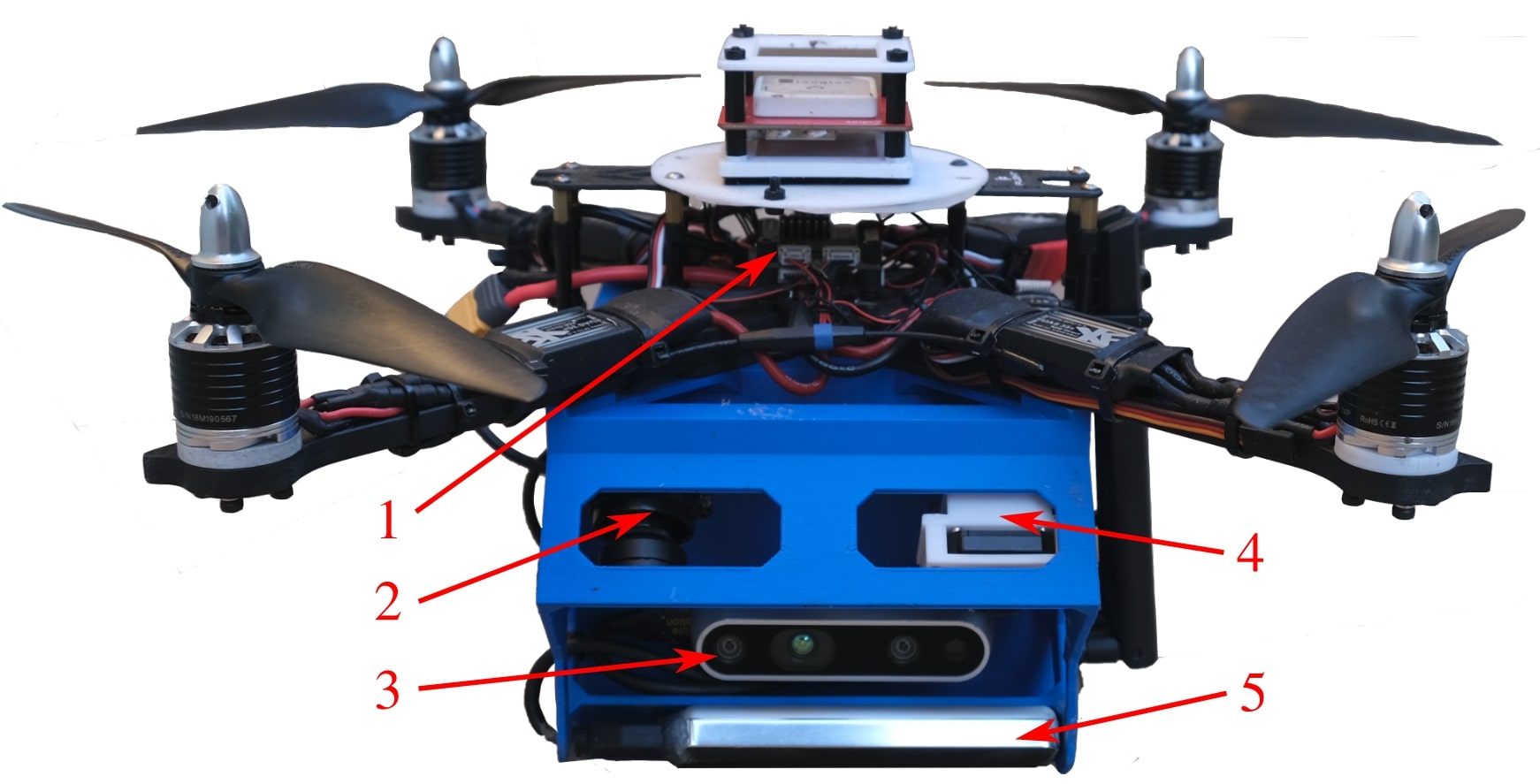}} \\
	\subfigure{\includegraphics[width=0.9\linewidth]{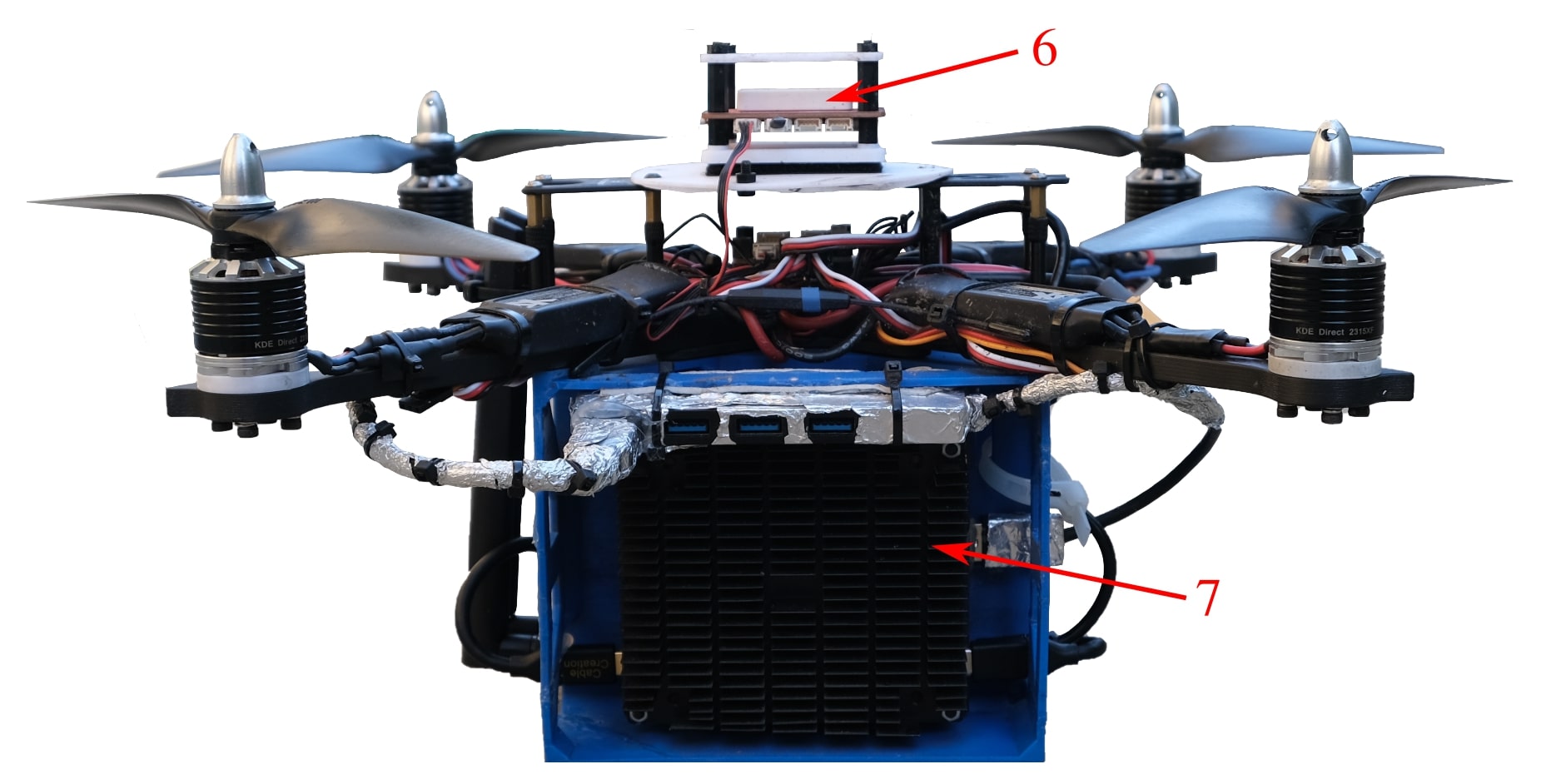}}
	\caption{The custom built quadcopter. 1 - Pixracer flight controller; 2 - RGB camera (not used in feedback); 3 - D435i depth camera; 4 - infra-red camera (not used in feedback); 5 - T265 tracking camera; 6 - GPS (not used in feedback); 7 - Jetson AGX Xavier
	}
	\label{fig:vehicle}
\end{figure}

\begin{figure}
	\begin{center}    \includegraphics[width = 0.95 \linewidth]{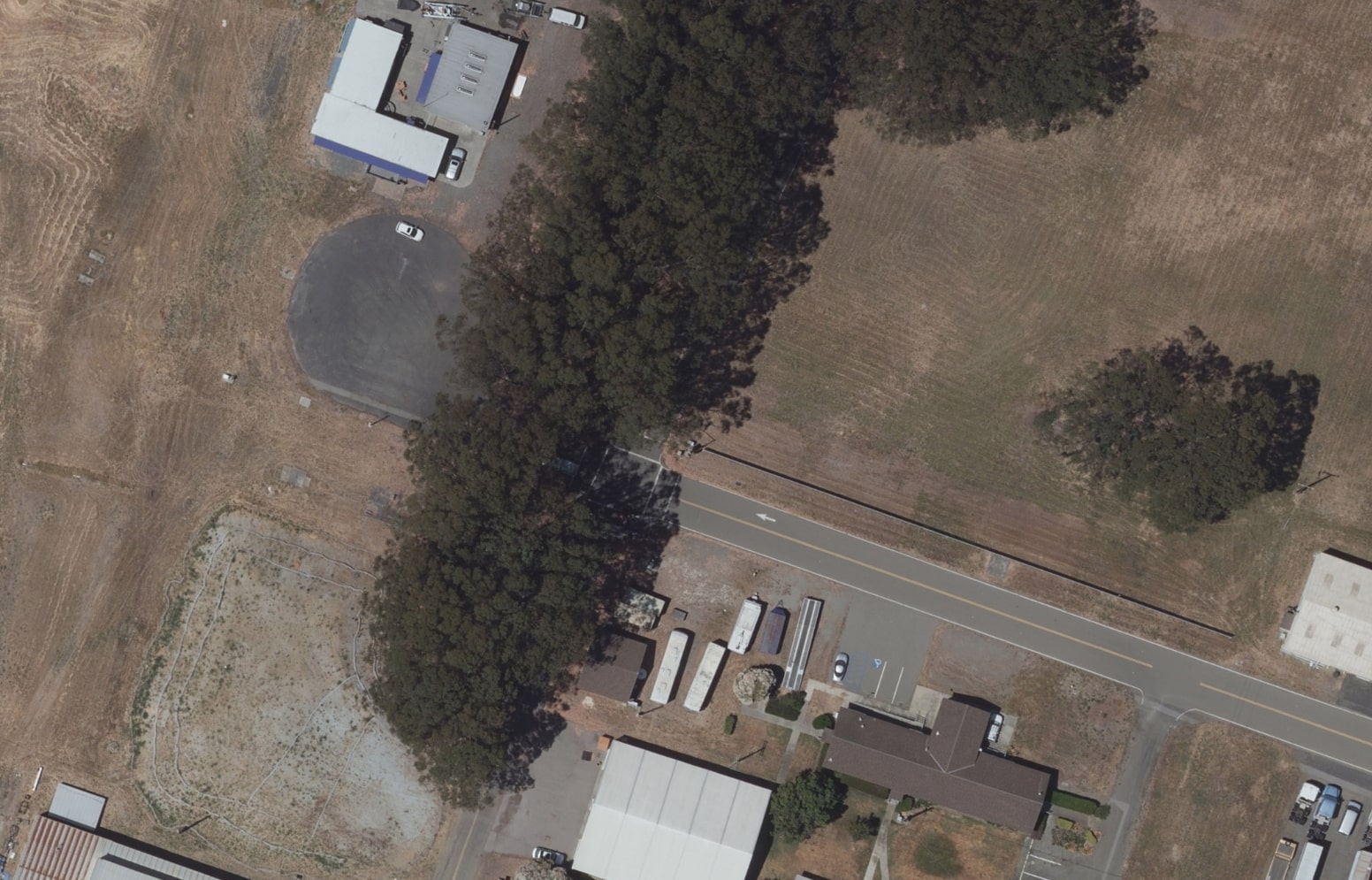}
	\end{center}
	\caption{Satellite image of the small forest at the Richmond Field Station where the experiment was conducted. Image is from {\tt www.usgs.gov} .}
	\label{fig:forest_satellite}
\end{figure}

\begin{figure*}
\includegraphics[width=0.9\textwidth]{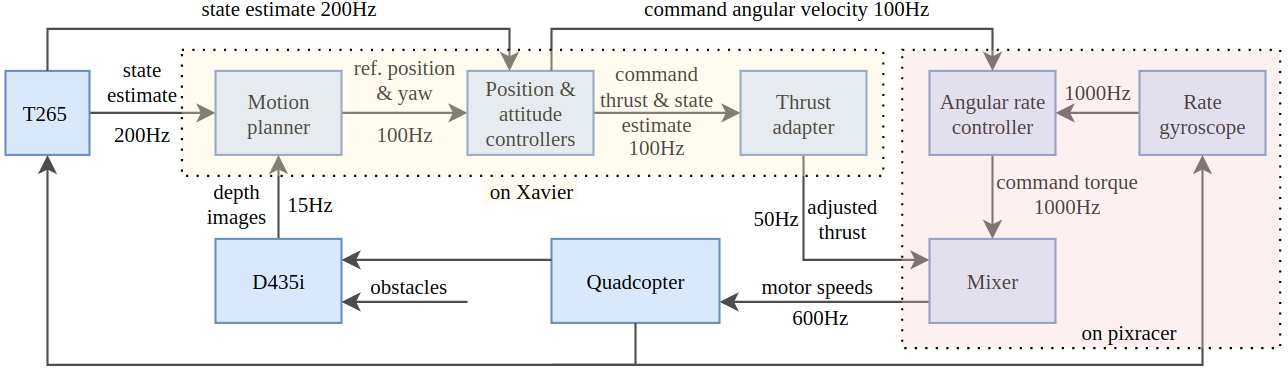}
\centering
\caption{A block diagram of the system, showing the relationship between components. The yellow shaded area contains components running on the AGX Xavier on-board computer, while the red shaded area contains components running on the Pixracer flight controller.
}
\label{fig:block-diagram}
\end{figure*}

We have shown in outdoor experiments that the system is able to navigate in a complex forest experiment with a maximum speed of 2.7 m/s. The experiment was repeated several times and we got similar performance. In this section we give detailed explanation of one of these experiments.
%%%%%%%%%%%%%%%%%%%%%%%%%%%%%%%%%%%%%%%%%%%%%%%%%%%%%%%%%%%%%%%%%%%%%%%%%%%%%%%%
\subsection{System setup}
A custom-built quadcopter, as shown in \figref{fig:vehicle}, was used through the experiments. It weighs 2.4 kg and the distance between two diagonal motors is 382 mm. The diameter of each propeller is 229 mm. On the vehicle, an Intel D435i depth camera is installed for collision avoidance and an Intel T265 camera is installed for state estimation. The depth camera is forward-looking to detect obstacles in forward flights, and the T265 camera has a 37-degree angle with the horizontal ground, to track features on the ground for state estimation and to avoid view occlusion from other parts of the vehicle. The GPS and two other cameras (one is a RGB camera and the other one is an infra-red camera) on the vehicle are not used for the vehicle's motion planning. The relationship between the components of the system, is shown in Fig. \ref{fig:block-diagram}. The RAPPIDS motion planner, the position and attitude controllers, and the thrust adapter runs on an on-board computer (Jetson AGX Xavier), at a frequency of 100Hz. The command thrust and angular velocity for each axis are then sent to the Pixracer flight controller via serial communication. The Pixracer runs the standard PX4 firmware and is used for the low-level control of the vehicle. The position controller is a PD controller, while the attitude and angular velocity controllers are P controllers.

%%%%%%%%%%%%%%%%%%%%%%%%%%%%%%%%%%%%%%%%%%%%%%%%%%%%%%%%%%%%%%%%%%%%%%%%%%%%%%%%
\subsection{Obstacle avoidance experiment}
The experiments were conducted at a small forest at the Richmond Field Station (37.915535 N, -122.335059 E), as shown in \figref{fig:forest_satellite}. The vehicle's radius $r$ (shown in Fig. \ref{gp:depth_image}) was set to 0.6m for the planner during the experiment, leaving a minimum safety margin of about 0.3 m between the vehicle and the nearest obstacles. The velocity constraint $v_{max}$ in section \ref{ssec:velocity-check} was set to 3 m/s because of the limit of the T265 tracking camera (a speed of above 3 m/s could make the state estimation of T265 unreliable), and trajectories exceeding this speed limit will be rejected.
\begin{figure}
	\begin{center}    \includegraphics[width = 0.95 \linewidth]{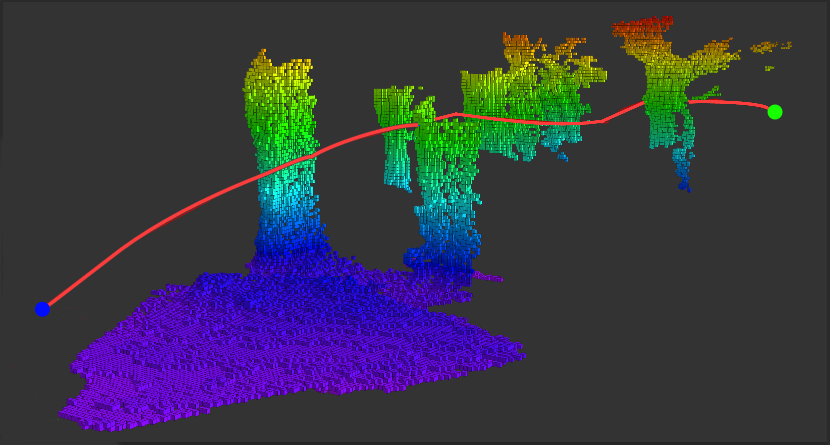}
	\end{center}
	\caption{Path of the vehicle (red line) during the autonomous collision avoidance flight from the start point (marked with a blue dot) to the end point (marked with a green dot). The trees detected by the depth camera were visualized using Octomap \cite{hornung_octomap_2013}. The distance between the start point and the end point is 30 meters. } 
	\label{fig:path}
\end{figure}

\begin{figure}
	\begin{center}    \includegraphics[width = 0.95 \linewidth]{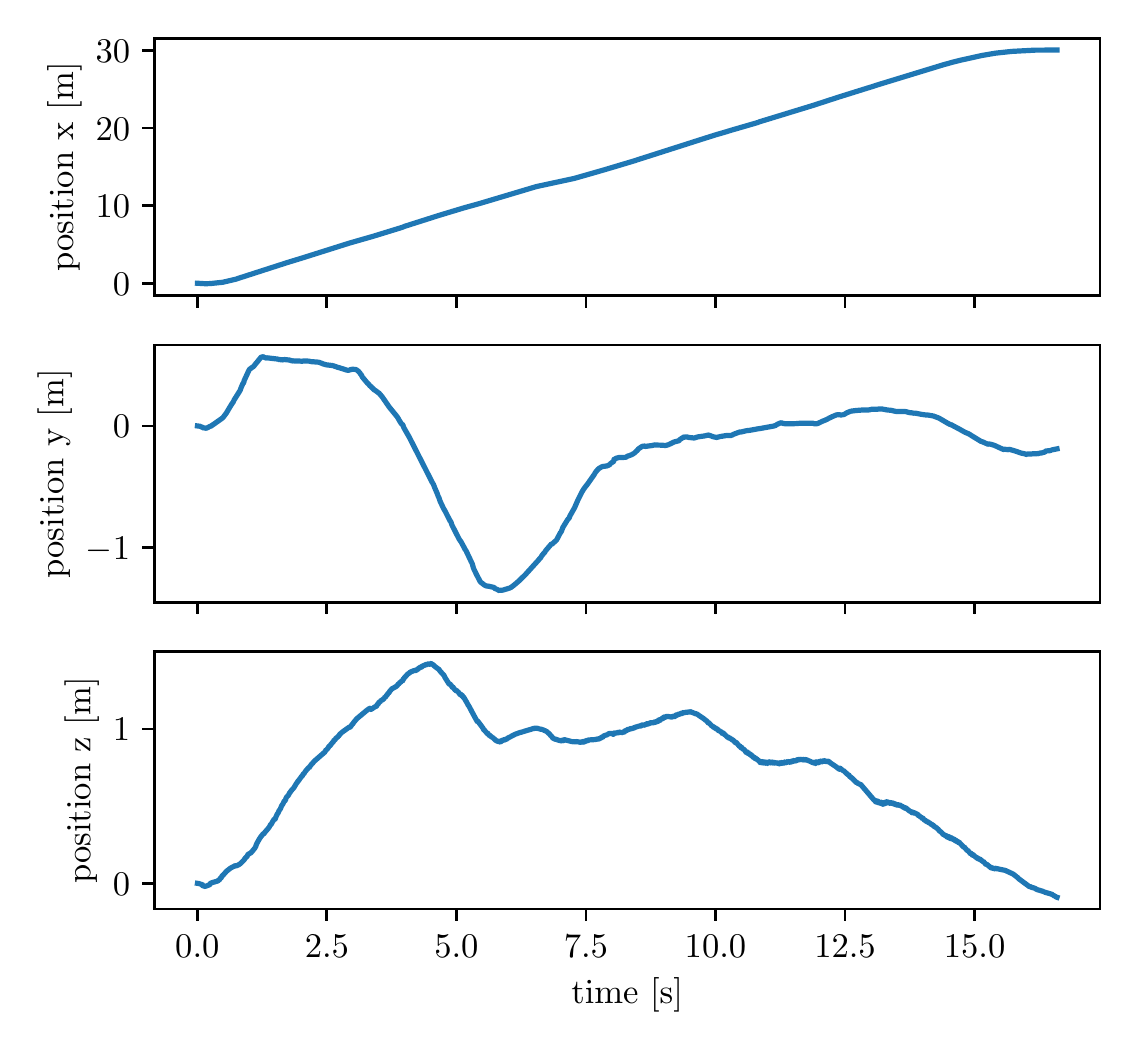}
	\end{center}
	\caption{The estimated position of the vehicle during the autonomous collision avoidance flight, with respect to the starting point of the autonomous flight. The target point of the vehicle was at x = 30 m, y = 0 m, z = 0 m. }
	\label{fig:path_position}
\end{figure}

\begin{figure}
	\begin{center}    \includegraphics[width = 0.95 \linewidth]{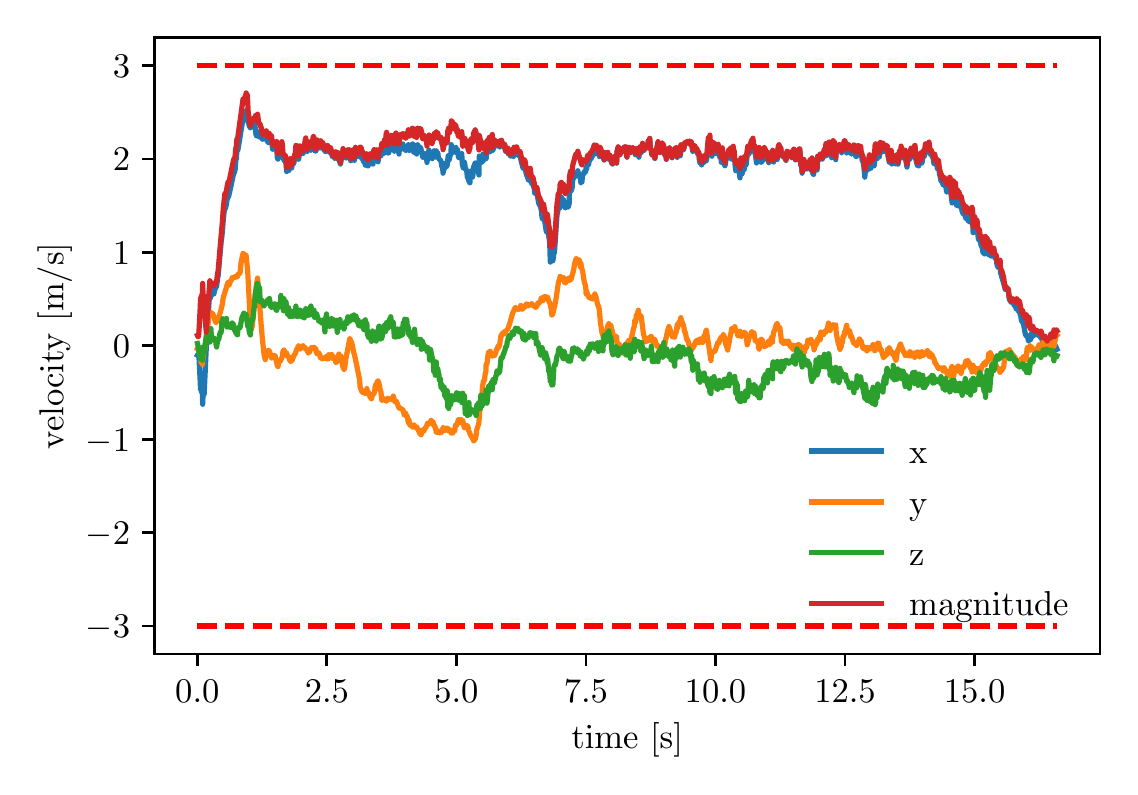}
	\end{center}
	\caption{The estimated velocity for each axis of the vehicle, as well as the Euclidean norm (magnitude) of the velocity during the autonomous collision avoidance flight. With the velocity check in section \ref{ssec:velocity-check}, the velocity on none of the three axis exceeds the velocity limit of 3.0 m/s (marked as red dashed lines).}
	\label{fig:traj_velocity}
\end{figure}

The vehicle was first controlled to take off manually to about 1 m above the ground and then switched to autonomous hovering at its current position, which was used as the starting point. The target point was set to be 30 meters forward with respect to the starting point of the vehicle, to fly the vehicle to the other side of the forest. 
When the vehicle was close to the target point (less than 1 m in this case), the motion planner in Fig. \ref{fig:block-diagram} stoped generating new trajectories, and the vehicle tracked the last reference trajectory to reach the target point. 
After the vehicle reached the target point, it hovered there and waited for the pilot to send other commands, e.g. landing. 
In the experiment the vehicle was able to reach the target point while generating and tracking collision-free trajectories. The path of the vehicle is visualized in \figref{fig:path} and its position is shown in \figref{fig:path_position}. The manual take-off and landing part are omitted and only the autonomous collision avoidance flight part is plotted for clarity. 
With the velocity check in section \ref{ssec:velocity-check} on the sampled trajectories, the velocity on none of the three axis exceeds the velocity limit of 3.0 m/s, shown in Fig. \ref{fig:traj_velocity}.

\begin{figure}
	\begin{center}    \includegraphics[width = 0.98 \linewidth]{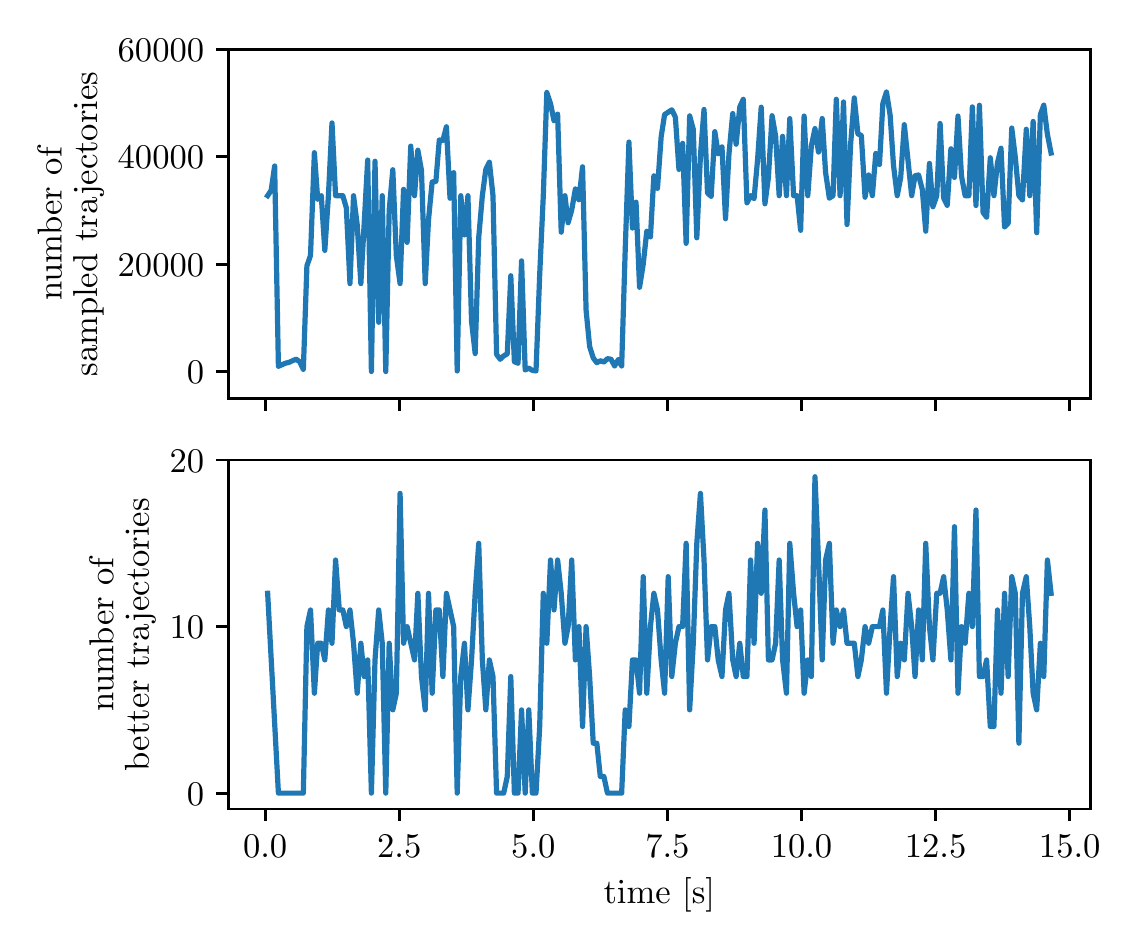}
	\end{center}
	\caption{A large number of trajectories were sampled during the autonomous flight of the vehicle, as shown in the first row. The sampled trajectories then went through the checks in Algorithm~\ref{alg:constraints-check}, and the number of trajectories that are better than the current trajectory is shown in the second row.}
	\label{fig:planner}
\end{figure}

The number of sampled trajectories and better-than-current trajectories (i.e. trajectories that pass all the checks in Algorithm~\ref{alg:constraints-check} and have a lower cost than the current reference trajectory) throughout the experiment is shown in \figref{fig:planner}. 
Thanks to the computational efficiency of the algorithm, a large number of sampled trajectories could be processed on-board in real-time.
The current reference trajectory was updated when a better-than-current trajectory was found, which happened most of the time during the flight. 
When no better trajectory was found (e.g. when the view of the depth camera was occluded by the obstacles), the vehicle would track the current trajectory.
After 14.7 s, the vehicle was within 1 m to the target point, and the trajectory generator stopped generating new trajectories and followed the last reference trajectory to reach the target point.

 %%%%%%%%%%%%%%%%%%%%%%%%%%%%%%%%%%%%%%%%%%%%%%%%%%%%%%%%%%%%%%%%%%%%%%%%%%%%%%%%

\section{CONCLUSIONS}
In this paper, we presented various considerations of the RAPPIDS motion planner for outdoor flights.
We introduced the velocity constraint of the planner to satisfy the speed limit of the visual-inertial odometry camera, increased the trajectory sampling efficiency based on the prior that sampling close to the edge of field of view of the depth camera is prone to result in in-collision trajectories, and used estimated acceleration instead of noisy IMU acceleration measurements.
In addition, a new utility function was proposed to consider not only maximizing the average velocity toward the goal but also keeping the vehicle around the goal.
A thrust adaptation method is introduced to compensate decrease in motor thrusts due to voltage drop during flights.
Lastly, the experimental results in a challenging outdoor environment were presented, which validated the ability of this system to autonomously navigate through complex obstacles outdoors.

\section*{Acknowledgement}
Research was sponsored by the Army Research Laboratory and was accomplished under Cooperative Agreement Number W911NF-20-2-0105. The views and conclusions contained in this document are those of the authors and should not be interpreted as representing the official policies, either expressed or implied, of the Army Research Laboratory or the U.S. Government. The U.S. Government is authorized to reproduce and distribute reprints for Government purposes notwithstanding any copyright notation herein.
The experimental testbed at the HiPeRLab is the result of contributions of many people, a full list of which can be found at \url{hiperlab.berkeley.edu/members/}.

%                                   % the textheight of the last page by a suitable amount.
%                                   % This command does not take effect until the next page
%                                   % so it should come on the page before the last. Make
%                                   % sure that you do not shorten the textheight too much.

% BIBLIOGRAPHY
\typeout{} 
{
\bibliographystyle{IEEEtran}
\bibliography{bibliography.bib}
}

\end{document}